\documentclass[12pt]{article}

\usepackage{graphicx}   % For figures
\usepackage{amsmath}    % Math equations
\usepackage{amssymb}    % Math symbols
\usepackage{hyperref}   % Hyperlinks
\usepackage{booktabs}   % Tables

\usepackage{algorithm}  % Algorithm environment
\usepackage{algpseudocode} % Algorithmic pseudocode

\usepackage{cite}       % Citation management
\usepackage{xcolor}     % Colors for highlighting
\usepackage{geometry}   % Page layout

\usepackage{subcaption}

\usepackage{tablefootnote}

\title{HERCULES: Hierarchical Embedding-based Recursive Clustering Using LLMs for Efficient Summarization}
\author{
	Gabor Petnehazi \\ 
	University of Debrecen \\
	\texttt{gabor.petnehazi@science.unideb.hu}
	\and
	Bernadett Aradi \\
	University of Debrecen \\
	\texttt{aradi.bernadett@inf.unideb.hu}
}
\date{}

\newcommand{\hercules}{\textsc{Hercules}}

\begin{document}
	
	\maketitle
	
	\begin{abstract}
		The explosive growth of complex datasets across various modalities necessitates advanced analytical tools that not only group data effectively but also provide human-understandable insights into the discovered structures. We introduce HERCULES (Hierarchical Embedding-based Recursive Clustering Using LLMs for Efficient Summarization), a novel algorithm and Python package designed for hierarchical k-means clustering of diverse data types, including text, images, and numeric data (processed one modality per run). HERCULES constructs a cluster hierarchy by recursively applying k-means clustering, starting from individual data points at level 0. A key innovation is its deep integration of Large Language Models (LLMs) to generate semantically rich titles and descriptions for clusters at each level of the hierarchy, significantly enhancing interpretability. The algorithm supports two main representation modes: `direct' mode, which clusters based on original data embeddings or scaled numeric features, and `description' mode, which clusters based on embeddings derived from LLM-generated summaries. Users can provide a `topic\_seed' to guide LLM-generated summaries towards specific themes. An interactive visualization tool facilitates thorough analysis and understanding of the clustering results. We demonstrate HERCULES's capabilities and discuss its potential for extracting meaningful, hierarchical knowledge from complex datasets.
	\end{abstract}
	
	
	\section{Introduction}
	\label{sec:introduction}
	Clustering, a fundamental task in unsupervised machine learning, addresses the challenge of discovering inherent structures in data without predefined labels. A key difficulty lies not only in the ill-posed nature of the task—where the definition of a ``good'' cluster depends on the specific analytical goal—but also in the subsequent, often Herculean, task of interpreting these abstract groupings to extract human-understandable insights.
	
	While traditional flat clustering algorithms partition data into a single set of groups, hierarchical methods \cite{jain1999data} offer a richer, multi-level perspective. They reveal nested structures at different granularities, providing a more flexible and comprehensive view of data organization.
	
	However, interpreting these hierarchies, especially for high-dimensional or non-numeric data like text and images, can be challenging. The sheer volume of items within clusters and the abstract nature of their representations often obscure the underlying themes. Recent advancements in Large Language Models (LLMs) \cite{brown2020language, touvron2023llama} have demonstrated remarkable capabilities in natural language understanding and generation, opening new avenues for enhancing data analysis pipelines.
	
	In this paper, we introduce \hercules{} (Hierarchical Embedding-based Recursive Clustering Using LLMs for Efficient Summarization), a novel algorithm for hierarchical data analysis. The open-source implementation of our method is available as the \texttt{pyhercules} package\footnote{The \texttt{pyhercules} source code is available at \url{https://github.com/bandeerun/pyhercules}}. \hercules{} performs hierarchical clustering by recursively applying the k-means algorithm. Its novelty lies in the deep integration of LLMs to generate concise, human-readable titles and descriptions for each cluster at every level of the hierarchy. This LLM-driven summarization aims to make the resulting cluster structure significantly more interpretable and actionable for users.
	
	\hercules{} is designed to be versatile, capable of processing text, image, or numeric datasets (one modality at a time per execution). It leverages embedding models to convert text and images into dense vector representations suitable for k-means. Numeric data is appropriately scaled. The algorithm offers two distinct representation strategies for clustering:
	\begin{enumerate}
		\item \textbf{Direct Mode:} Clusters are formed based on the embeddings of the original data items (or scaled numeric features for numeric data).
		\item \textbf{Description Mode:} Clusters are formed based on the embeddings of the LLM-generated textual descriptions of the (sub-)clusters from the previous level. This mode allows the semantic understanding of the LLM to directly influence the clustering process at higher levels.
	\end{enumerate}
	
	An optional \texttt{topic\_seed} can be provided to steer the LLM's summarization towards a particular theme or context, making the descriptions more relevant to a specific analytical goal. The Python package offers robust tools for evaluating cluster quality. To facilitate exploration of the results, it is complemented by a separate, standalone interactive visualization application that consumes the output of the package.
	
	The main contributions of this work are:
	\begin{itemize}
		\item A novel hierarchical clustering algorithm, \hercules{}, that synergistically combines recursive k-means with LLM-based summarization for enhanced interpretability.
		\item Support for flexible uni-modal clustering of text, image, and numeric data, with configurable representation modes.
		\item An adaptive LLM prompting strategy that incorporates representative samples from underlying data, child cluster summaries (for deeper levels), and relevant numeric statistics to generate informative cluster descriptions.
		\item Option for a \texttt{topic\_seed} for guided summarization.
		\item A comprehensive Python package with built-in evaluation metrics, complemented by a standalone application for interactive visualization.
	\end{itemize}
	
	The remainder of this paper is organized as follows: Section~\ref{sec:related_work} discusses related research. Section~\ref{sec:hercules_algorithm} details the \hercules{} algorithm. Section~\ref{sec:pseudocode} presents the algorithm pseudocode. Section~\ref{sec:package_features} describes the features of the Python package and visualization application. Section~\ref{sec:experimental_setup} outlines the experimental setup, followed by a discussion of results in Section~\ref{sec:results_discussion}. Finally, Section~\ref{sec:conclusion} concludes the paper and suggests future work.
	
	\section{Related Work}
	\label{sec:related_work}
	The development of \hercules{} intersects with and aims to extend several key research areas: hierarchical clustering methodologies, techniques for enhancing cluster interpretability, and the burgeoning field of applying Large Language Models (LLMs) to clustering tasks.
	
	The foundational clustering mechanism in \hercules{} is adapted from Vo et al.\ (2024)~\cite{vo2024automatic}. We adopt their method of constructing a hierarchy by recursively applying k-means to the centroids of the preceding level's clusters. Furthermore, our optional iterative resampling loop for centroid refinement is also based on the technique described in their work. Vo et al.\ utilize this hierarchical structure and resampling process for automatic data curation in self-supervised learning, aiming to rebalance large datasets by achieving a more uniform distribution of cluster centroids. While \hercules{} employs this same core mechanism for building the cluster hierarchy, its primary innovation and focus diverge significantly. \hercules{} is designed as a general-purpose, user-facing hierarchical clustering tool with a central emphasis on interpreting the discovered structure through deep and systematic integration of LLM-based summarization at each level. The novelty of our work is not in the clustering algorithm itself, but in how we augment it with LLMs to create an interactive and interpretable data exploration system.
	
	Recent advancements in LLMs have shown remarkable capabilities in text summarization, abstraction, and even direct application in clustering tasks. Viswanathan et al.\ (2024) explored few-shot semi-supervised text clustering by incorporating LLMs at three stages: improving input features, providing pairwise constraints, and post-correction~\cite{viswanathan2024large}. Huang and He (2024) propose a framework that transforms text clustering into a classification task using LLMs~\cite{huang2024text}. Tipirneni et al.\ (2024) (CACTUS) focus on supervised clustering of entity subsets using an LLM to capture context~\cite{tipirneni2024context}. Miller and Alexander (2025) demonstrated that LLM-generated embeddings for short text yield more distinctive and human-interpretable clusters than doc2vec or latent Dirichlet allocation~\cite{miller2025human}. The LLooM algorithm uses LLMs to extract high-level concepts from text data~\cite{lam2024concept}. Tamkin et al.\ (2024) (Clio) use k-means and LLMs to build hierarchies of clusters to reveal patterns in AI assistant usage~\cite{tamkin2024clio}. Kwon et al.\ (2023) introduce a novel methodology that leverages vision-language models and large language models to perform image clustering based on user-specified, iteratively refined text criteria~\cite{kwon2023image}. These works underscore the growing trend of integrating LLMs to not only perform clustering but also to make the results more understandable and aligned with human intuition. While these works highlight various ways LLMs can aid clustering, \hercules{} provides a distinct unsupervised, recursive framework where LLM interactions are systematized for hierarchical summarization and, optionally, for guiding the clustering process itself through the `description' mode. \hercules{}'s adaptive prompting strategy aims to generate rich, context-aware descriptions across various data types within its hierarchical structure.
	
	The use of pre-trained embedding models (e.g., BERT \cite{devlin2019bert}, SBERT \cite{reimers2019sentence}, CLIP \cite{radford2021learning}) is a common precursor to clustering text and image data. \hercules{} leverages this common practice for its `direct' mode and also uses embeddings of LLM-generated descriptions in its `description' mode, effectively bridging the gap between raw data representations and higher-level semantic interpretations provided by LLMs.
	
	Making complex clustering results accessible often requires interactive visualization tools. Systems like the LLooM Workbench~\cite{lam2024concept} and the Clio interface~\cite{tamkin2024clio} allow users to dynamically explore cluster hierarchies and inspect cluster contents. In a similar vein, \hercules{} is complemented by a standalone visualization application.
	
	In summary, \hercules{} integrates a robust hierarchical k-means algorithm with iterative resampling, as introduced by Vo et al.\ (2024), and builds upon this foundation with a novel system for LLM-driven summarization and interpretation. It extends concepts from hierarchical clustering, text representation, and LLM-in-the-loop paradigms to offer a versatile system for generating and understanding multi-level data structures, complemented by an interactive visualization tool.

	\section{The \hercules{} Algorithm}
	\label{sec:hercules_algorithm}
	\hercules{} constructs a hierarchy of clusters by iteratively applying k-means. Each cluster in the hierarchy is represented by a \texttt{Cluster} object, storing its level, constituent items (children or original data points), LLM-generated title and description, and its vector representation. The core logic is shown in Algorithms~\ref{alg:hercules_main}-\ref{alg:hercules_loop}, in Section~\ref{sec:pseudocode}.
	
	\subsection{Input Data Preparation and Level 0 Initialization}
	\label{ssec:input_prep_l0}
	\hercules{} accepts various input data formats:
	\begin{itemize}
		\item \textbf{Text Data:} A list of strings or a dictionary/Pandas Series of texts.
		\item \textbf{Image Data:} A list of image file paths or identifiers that an image embedding/captioning client can process. Image type is heuristically detected based on file extensions.
		\item \textbf{Numeric Data:} A list of numbers, list of numeric vectors, a dictionary of numeric vectors, or a Pandas DataFrame/Series. Numeric data is standardized (zero mean, unit variance) using \texttt{sklearn.preprocessing.StandardScaler}. The original, unscaled numeric data is preserved for statistical computations used in LLM prompts.
	\end{itemize}
	The algorithm automatically detects the input data type. For numeric data, users can provide optional metadata (e.g., variable names, units, descriptions) via a \texttt{numeric\_metadata} dictionary, which is used to enrich LLM prompts for numeric clusters.
	
	Level 0 clusters represent the individual input data items. For each item, a \texttt{Cluster} object is created with \texttt{level=0}.
	\begin{itemize}
		\item \textbf{Initial Titles:} A preliminary title is extracted. For text, it's often the first sentence or first few words. For images, it's derived from the filename. For numeric data, it might reference the original ID.
		\item \textbf{Initial Descriptions and Representations:}
		\begin{sloppypar}
			\begin{itemize}
				\item If \texttt{use\_llm\_for\_l0\_descriptions} is true for text/numeric data, or if an \texttt{image\_captioning\_client} is provided for images, \hercules{} employs these to generate initial descriptions. The prompts for L0 LLM summarization (if used) are simpler, typically focusing on the single item's content.
				\item Otherwise, for text, the description is a truncated snippet of the original text. For numeric data, it's a string representation of values. For images without captioning, it's a placeholder.
			\end{itemize}
    	\end{sloppypar}
	\end{itemize}
	These initial titles and descriptions (or captions) form the basis for the \texttt{description} mode representation at Level 0 if that mode is selected. This initialization process is formally detailed in Algorithm~\ref{alg:hercules_l0}.
	
	\subsection{Representation Generation}
	\label{ssec:representation_generation}
	At each level $i \ge 0$, clusters (or items at L0) need a vector representation for the (k-means) clustering at level $i+1$. \hercules{} supports two modes:
	
	\begin{enumerate}
		\item \textbf{Direct Mode}:
		\begin{itemize}
			\item For L0:
			\begin{sloppypar}
				\begin{itemize}
					\item Text: Embeddings of the original text items using the provided \texttt{text\_embedding\_client}.
					\item Image: Embeddings of the original images using the \texttt{image\_embedding\_client}.
					\item Numeric: The scaled numeric feature vectors.
				\end{itemize}
			\end{sloppypar}
			\item For $L_i, i \ge 1$: The representation of a cluster is the centroid of its children's L$_{i-1}$ representations (which are themselves centroids or L0 direct representations). The vector space (e.g., `embedding' or `numeric') is inherited.
		\end{itemize}
		
		\item \textbf{Description Mode}:
		\begin{itemize}
			\item For L0: Embeddings of the L0 descriptions/captions (generated as in Section~\ref{ssec:input_prep_l0}) using the \texttt{text\_embedding\_client}.
			\item For $L_i, i \ge 1$: Embeddings of the LLM-generated descriptions of these $L_i$ clusters using the \texttt{text\_embedding\_client}.
		\end{itemize}
	\end{enumerate}
	\begin{sloppypar}
		A \texttt{Cluster} object stores its \texttt{representation\_vector} (centroid) and, separately, its \texttt{description\_embedding} if available.
	\end{sloppypar}
	
	\subsection{Hierarchical K-Means Clustering}
	\label{ssec:hierarchical_kmeans}
	The core clustering process in \hercules{}, detailed in Algorithm~\ref{alg:hercules_loop}, is adapted from the hierarchical k-means algorithm proposed by Vo et al.\ (2024)~\cite{vo2024automatic}. Starting with the representations of Level 0 clusters (or Level $i-1$ clusters for $i \ge 1$), it performs the following steps recursively:
	\begin{enumerate}
		\item \textbf{Vector Collection:} Collect valid vectors for clustering (\texttt{representation\_vector}s in case of direct mode, \texttt{description\_embedding}s in case of description mode) from the current level's clusters. If insufficient vectors are available (e.g., less than \texttt{min\_clusters\_per\_level}), the hierarchy stops.
		\item \textbf{Determine $k$ (Number of Clusters):}
		\begin{itemize}
			\item \textbf{Fixed $k$:} If \texttt{level\_cluster\_counts} is provided, the $k$ for the current level is taken from this list.
			\item \textbf{Automatic $k$:} If \texttt{level\_cluster\_counts} is \texttt{None}, $k$ is determined by optimizing a chosen metric.
		\end{itemize}
		The effective $k$ is always $\ge \text{min\_clusters\_per\_level}$ and less than the number of items being clustered. If $k=1$ is chosen, the hierarchy stops.
		\item \textbf{K-Means Clustering:} Apply \texttt{sklearn.cluster.KMeans} with the determined $k$ to the collected vectors. 
		
		Additionally, \hercules{} incorporates the optional iterative refinement process described by Vo et al.\ (2024)~\cite{vo2024automatic}, enabled by the \texttt{use\_resampling} parameter (see lines 12-22 in Algorithm~\ref{alg:hercules_loop}). If `True', after an initial k-means pass, the algorithm performs the following refinement loop for a specified number of iterations:
		\begin{enumerate}
			\item \textbf{Resample Core Members:} For each of the $k$ clusters, it identifies the subset of points closest to that cluster's current centroid. These ``core'' members are collected into a new, smaller dataset, $R$.
			\item \textbf{Refine Centroids:} K-means is run again, but only on the resampled dataset $R$. This step calculates new, more robust centroids that are less influenced by potential outliers on the periphery of clusters.
			\item \textbf{Re-assign Full Dataset:} The entire original set of input vectors for the level is re-assigned to these newly refined centroids, producing an updated set of cluster labels.
		\end{enumerate}
		
		\item \textbf{Create New Level Clusters:} For each of the $k$ groups identified in the final assignment, a new parent \texttt{Cluster} object is created at the next higher level. The children of this new parent are the clusters from the previous level assigned to this group by k-means. The \texttt{representation\_vector} of the new parent cluster is set to the final centroid computed by k-means.
	\end{enumerate}
	This process repeats, building the hierarchy level by level, until a stopping condition is met (e.g., maximum configured levels, $k=1$, or insufficient items to cluster). The highest level achieved is stored.
	
\subsection{LLM-Powered Cluster Summarization}
\label{ssec:llm_summarization}
A core innovation of \hercules{} is the generation of interpretable summaries. After new clusters are formed at level $i \ge 1$ (or for L0 items if configured), \hercules{} generates their titles and descriptions using the provided \texttt{llm\_client} (see lines 31-32 in Algorithm~\ref{alg:hercules_loop}). This is orchestrated through a sophisticated, dynamically constructed prompt.

\textbf{Prompt Construction:} For each cluster to be summarized, a detailed prompt is constructed. The prompt is a carefully engineered set of instructions that asks the LLM to return a concise title (e.g., max 5-7 words) and a short description (e.g., 1-2 sentences) in a structured JSON format. To provide the LLM with sufficient context to generate a high-quality summary, the prompt dynamically includes several key pieces of information:
\begin{sloppypar}
	\begin{itemize}
		\item \textbf{Cluster Metadata:} Level, number of L0 descendants, original data type.
		\item \textbf{Representative L0 Samples:} A configurable number (\texttt{prompt\_l0\_sample\_size}) of L0 descendant items are sampled. Strategies include \texttt{centroid\_closest} (items whose L0 representations are closest to the current cluster's L$_{i}$ representation vector) or \texttt{random}. Text samples are truncated (\texttt{prompt\_l0\_sample\_trunc\_len}). Numeric samples display values with specified precision and count (\texttt{prompt\_l0\_numeric\_repr\_max\_vals}, \texttt{prompt\_l0\_numeric\_repr\_precision}).
		\item \textbf{Immediate Child Summaries (for $L_i, i \ge 2$):} If \texttt{prompt\_include\_immediate\_children} is true, summaries (titles and truncated descriptions) of a sample of immediate child clusters (from level $i-1$) are included. This gives the LLM context about the sub-structure. Sampling strategies (\texttt{prompt\_immediate\_child\_sample\_strategy}) and size (\texttt{prompt\_immediate\_child\_sample\_size}) are configurable.
		\item \textbf{Numeric Statistics (for numeric data):} If the original data is numeric, key statistics (mean, min, max, std) are computed from the \textit{original, unscaled} L0 data points within the cluster. These are presented along with variable names and any provided metadata (units, descriptions). The number of variables shown (\texttt{prompt\_max\_stats\_vars}) and precision (\texttt{prompt\_numeric\_stats\_precision}) are configurable.
		\item \textbf{Topic Seed:} If a \texttt{topic\_seed} is provided to \hercules{}, it's included in the prompt to guide the LLM's focus.
	\end{itemize}
\end{sloppypar}
An annotated example of the complete prompt template is provided in Appendix~\ref{app:prompt_template}. Token count for prompts is estimated, and content is managed to stay within LLM context limits (\texttt{max\_prompt\_tokens}).

\textbf{Batched LLM Calls and Parsing:}
To handle potentially many clusters, \hercules{} batches requests to the LLM.
\begin{sloppypar}
	\begin{itemize}
		\item \textbf{Adaptive Batching:} Starts with \texttt{llm\_initial\_batch\_size} and reduces it by \texttt{llm\_batch\_reduction\_factor} if prompts are too large or calls fail, down to \texttt{llm\_min\_batch\_size}.
		\item \textbf{Retries:} Failed LLM calls (e.g., due to API errors or timeouts) are retried up to \texttt{llm\_retries} times with an increasing \texttt{llm\_retry\_delay}.
		\item \textbf{Response Parsing:} LLM responses are expected in JSON format. \hercules{} parses this JSON, extracting titles and descriptions for each cluster ID in the batch. It handles common issues like markdown fences around JSON.
	\end{itemize}
\end{sloppypar}
The generated titles and descriptions are stored in the respective \texttt{Cluster} objects. If LLM summarization fails persistently for a cluster, a default failure message is used.
	
	\subsection{Topic-Guided Clustering}
	\label{ssec:topic_guidance}
	The optional \texttt{topic\_seed} parameter allows users to influence the LLM-generated summaries. By providing a specific theme or topic, the LLM is encouraged to frame its descriptions in that context. This can be particularly useful when analyzing data with a specific research question or business objective in mind, making the resulting hierarchy more aligned with the user's focus. While it primarily affects summarization, in \texttt{description} mode, this guidance can indirectly influence the clustering of higher levels.
	
	\subsection{Dimensionality Reduction}
	\label{ssec:dimensionality_reduction}
	\hercules{} supports dimensionality reduction of representation vectors, primarily for visualization. Currently, Principal Component Analysis (PCA) is implemented.
	\begin{sloppypar}
		\begin{itemize}
			\item Reducers (e.g., PCA instances) are trained separately for `numeric' and `embedding' vector spaces using the L0 representations of the respective type.
			\item The number of components (\texttt{n\_reduction\_components}, typically 2 or 3 for visualization) is configurable.
			\item Once trained, these reducers can be applied to the \texttt{representation\_vector} and \texttt{description\_embedding} of any cluster in the hierarchy. The reduced vectors are stored in \texttt{representation\_vector\_reductions} and \texttt{description\_embedding\_reductions} dictionaries within the \texttt{Cluster} object.
		\end{itemize}
	\end{sloppypar}
	This feature aids in creating 2D or 3D scatter plots of clusters, colored by their assignments, which can be valuable for visual inspection.
	
	\subsection{Evaluation Capabilities}
	\label{ssec:evaluation_capabilities}
	The \hercules{} Python package includes an \texttt{evaluate\_level} method for assessing clustering quality at a specific hierarchy level. This method computes:
	\begin{sloppypar}
		\begin{itemize}
			\item \textbf{Traditional Internal Metrics:} Silhouette Score, Davies-Bouldin Index, and Calinski-Harabasz Index can be calculated based on the L0 item representations and their assignments to clusters at the specified level, or based on the (L--1) cluster representations and their assignment to L clusters.
			\item \textbf{LLM-based Internal Metrics:} If enabled (\texttt{calculate\_llm\_internal\_metrics=True}), the same internal metrics (Silhouette, Davies-Bouldin, Calinski-Harabasz) are calculated using the LLM-generated \textit{description embeddings} of the clusters at the specified level (or L0 item descriptions if evaluating L1 clusters based on L0 descriptions). This provides an assessment of cluster coherence based on their semantic summaries.
			\item \textbf{Topic Alignment Score:} If a \texttt{topic\_seed} was provided, this metric calculates the average cosine similarity between the topic seed's embedding and the description embeddings of the clusters at the evaluated level. This indicates how well the generated cluster themes align with the intended topic.
			\item \textbf{External Metrics (if ground truth provided):} Adjusted Rand Score, Normalized Mutual Information, Homogeneity, Completeness, and V-measure are computed if ground truth labels for L0 items are supplied.
		\end{itemize}
	\end{sloppypar}
	These evaluation metrics offer a quantitative way to understand the quality and characteristics of the generated hierarchical structure from different perspectives.
	
	\section{Algorithm Pseudocode}
	\label{sec:pseudocode}
	This section provides the detailed pseudocode for the \hercules{} algorithm. Algorithm~\ref{alg:hercules_main} shows the main procedure, Algorithm~\ref{alg:hercules_l0} details the initialization of leaf nodes, and Algorithm~\ref{alg:hercules_loop} describes the core iterative clustering and summarization loop.
	
	% =====================================================================
	% Algorithm 1: Main HERCULES Procedure
	% =====================================================================
	\begin{algorithm}[H]
		\caption{HERCULES (Main Procedure)}
		\label{alg:hercules_main}
		\begin{algorithmic}[1]
			\Require
			$D$: Input dataset (text, numeric, or image)
			\Statex \hspace{\algorithmicindent} $k_{\text{config}}$: Cluster counts per level (list or `auto')
			\Statex \hspace{\algorithmicindent} $repr\_mode$: Representation mode (`direct' or `description')
			\Statex \hspace{\algorithmicindent} Clients: $E_{\text{text}}, E_{\text{image}}, L, C_{\text{image}}$
			\Statex \hspace{\algorithmicindent} Resampling config: $use\_resampling, m, r_t$
			\Ensure
			$C_{\text{top}}$: A set of top-level clusters forming a hierarchy.
			
			\Procedure{Hercules}{$D, k_{\text{config}}, \dots$}
			\State $C_0 \gets \text{InitializeLevel0}(D, repr\_mode, L, C_{\text{image}}, E_{\text{text}}, E_{\text{image}})$
			\State $C_{\text{top}} \gets \text{HierarchicalLoop}(C_0, k_{\text{config}}, repr\_mode, L, E_{\text{text}}, use\_resampling, m, r_t)$
			\State \textbf{return} $C_{\text{top}}$
			\EndProcedure
		\end{algorithmic}
	\end{algorithm}

	% =====================================================================
	% Algorithm 2: Level 0 Initialization
	% =====================================================================
	\begin{algorithm}[H]
		\caption{Level 0 Initialization}
		\label{alg:hercules_l0}
		\begin{algorithmic}[1]
			\Procedure{InitializeLevel0}{$D, repr\_mode, L, C_{\text{image}}, E_{\text{text}}, E_{\text{image}}$}
			\State $D_{\text{std}}, D_{\text{orig\_num}} \gets \text{PrepareInput}(D)$ \Comment{Scale numerics, handle formats}
			\State $C_0 \gets \{\text{New Cluster}(d_i) \text{ for } d_i \in D_{\text{std}}\}$ \Comment{Create L0 leaf nodes}
			
			\Statex \Comment{Generate initial L0 descriptions and their embeddings}
			\State Run LLM, captioning, or snippet extraction to create $c.\text{description}$ for all $c \in C_0$
			\State $\{\text{desc\_emb}_i\} \gets E_{\text{text}}(\{c.\text{description} \text{ for } c \in C_0\})$
			\State Assign $c_i.\text{desc\_embedding} \gets \text{desc\_emb}_i$ for each $c_i \in C_0$
			
			\Statex \Comment{Generate L0 representation vectors for clustering}
			\If{$repr\_mode == \text{`direct'}$}
			\State $I_0 \gets \text{GetDirectVectors}(D_{\text{std}}, E_{\text{text}}, E_{\text{image}})$ \Comment{Embed/use raw data}
			\State Assign $c_i.\text{rep\_vector} \gets I_{0,i}$ for each $c_i \in C_0$
			\Else \Comment{$repr\_mode == \text{`description'}$}
			\State Assign $c.\text{rep\_vector} \gets c.\text{desc\_embedding}$ for each $c \in C_0$
			\EndIf
			\State \textbf{return} $C_0$
			\EndProcedure
		\end{algorithmic}
	\end{algorithm}

	% =====================================================================
	% Algorithm 3: Hierarchical Clustering Loop
	% =====================================================================
	\begin{algorithm}[H]
		\caption{Hierarchical Clustering Loop}
		\label{alg:hercules_loop}
		\begin{algorithmic}[1]
			\Procedure{HierarchicalLoop}{$C_0, k_{\text{config}}, repr\_mode, L, E_{\text{text}}, use\_resampling, m, r_t$}
			\State $C_{\text{current}} \gets C_0$
			\While{$|C_{\text{current}}| > \text{min\_clusters}$}
			\If{$repr\_mode == \text{`direct'}$}
			\State $I \gets \{c.\text{rep\_vector} \text{ for } c \in C_{\text{current}}\}$
			\Else \Comment{$repr\_mode == \text{`description'}$}
			\State $I \gets \{c.\text{desc\_embedding} \text{ for } c \in C_{\text{current}}\}$
			\EndIf
			
			\State $k \gets \text{DetermineK}(I, k_{\text{config}})$
			\If{$k \le 1$} \textbf{break} \EndIf
			
			\Statex \textit{// K-Means with optional iterative resampling}
			\State $L, C_{\text{centroids}} \gets \text{KMeans}(I, k)$ \Comment{Initial clustering of input $I$}
			\If{$use\_resampling$}
			\For{$s=1$ to $m$} \Comment{Resampling iterations}
			\State $R \gets \emptyset$ \Comment{Initialize resampled set}
			\For{$j=1$ to $k$}
			\State $I_{j} \gets \{v \in I \mid \text{label}(v) = j\}$
			\State $R_j \gets \{r_t \text{ closest points in } I_{j} \text{ to } C_{\text{centroids}}[j]\}$
			\State $R \gets R \cup R_j$
			\EndFor
			\State $\_, C_{\text{centroids}} \gets \text{KMeans}(R, k)$ \Comment{Update centroids on resampled data $R$}
			\State $L \gets \text{Assign}(I, C_{\text{centroids}})$ \Comment{Update labels on original data $I$}
			\EndFor
			\EndIf
			
			\Statex \textit{// Create and summarize new parent clusters}
			\State $C_{\text{next}} \gets \emptyset$
			\For{$j=1$ to $k$}
			\State $c_{\text{new}} \gets \text{New Cluster with centroid } C_{\text{centroids}}[j]$
			\State $c_{\text{new}}.\text{children} \gets \{c \in C_{\text{current}} \mid \text{label}(c, L) = j\}$
			\State $C_{\text{next}} \gets C_{\text{next}} \cup \{c_{\text{new}}\}$
			\EndFor
			
			\State Summarize $C_{\text{next}}$ using LLM client $L$ to generate titles/descriptions
			\State Generate and assign description embeddings for $C_{\text{next}}$ using $E_{\text{text}}$
			\State $C_{\text{current}} \gets C_{\text{next}}$
			\EndWhile
			\State \textbf{return} $C_{\text{current}}$
			\EndProcedure
		\end{algorithmic}
	\end{algorithm}
	
	\section{The \texttt{pyhercules} Python Package and Visualization Application}
	\label{sec:package_features}
	The \hercules{} algorithm is implemented and provided as a user-friendly Python package, available for installation as \texttt{pyhercules}. This facilitates its integration into data analysis workflows. 
	
	\subsection{Core Package Features}
	The package encapsulates the algorithm described in Section~\ref{sec:hercules_algorithm}. Key features include:
	\begin{itemize}
		\item \textbf{Hercules Class:} The main class to instantiate and run the clustering process. It takes various configuration parameters, including embedding and LLM client functions.
		\item \textbf{Cluster Class:} Represents individual clusters within the hierarchy, storing all relevant information.
		\item \textbf{Flexible Configuration:} Extensive options for representation mode, $k$ determination, LLM prompt customization, client functions (text/image embedding, LLM, image captioning), and more.
		\item \textbf{Data Handling:} Automatic detection of input data type (text, image, numeric) and preprocessing (e.g., scaling for numeric data).
		\item \textbf{Output and Serialization:}
		\begin{itemize}
			\item Methods to retrieve the cluster hierarchy (e.g., top-level clusters, all clusters).
			\item \texttt{print\_hierarchy()} for a simple text-based tree view.
			\item \texttt{get\_cluster\_membership\_dataframe()} to get a Pandas DataFrame showing L0 item membership across levels.
			\item \texttt{save\_model()} and \texttt{load\_model()} for serializing the \hercules{} object (configuration, state, and cluster data) to/from JSON, allowing results to be saved and reloaded.
		\end{itemize}
		\item \textbf{Evaluation Module:} The \texttt{evaluate\_level} method provides comprehensive metrics as described in Section~\ref{ssec:evaluation_capabilities}.
		\item \textbf{Logging and Details:} Option to save run details, including LLM prompts and responses, for debugging and transparency.
	\end{itemize}
	
	\subsection{Interactive Visualization Application}
	\label{ssec:dash_visualization_app}
	Beyond the core Python package capabilities, a separate, standalone Dash \cite{dash} web application is provided for enhanced interactive visualization and exploration of \hercules{} clustering results. This application is not part of the algorithm itself but it can run \hercules{} and then visualize. Its main features include:
	
	\begin{itemize}
		\item \textbf{Data Upload and Configuration Interface:} Users can upload their data files (text, images, or tabular data for numeric clustering) and ground truth labels directly through the web interface. They can also configure all major \hercules{} parameters, such as representation mode, cluster counts (or auto-k settings), choice of embedding models and LLMs (from a predefined list if \texttt{pyhercules\_functions.py} is set up), topic seed, and other algorithm-specific settings.
		\item \textbf{Run Hercules Algorithm:} The application can execute the \hercules{} clustering algorithm based on the uploaded data and user configurations.
		\item \textbf{Interactive Scatter Plot Visualization:}
		\begin{itemize}
			\item Displays clusters (or L0 items if their count is below a threshold) as points in a 2D space, typically derived from PCA reduction of their representation vectors (either direct or description-based embeddings).
			\item Points are colored by their cluster ID at a selected hierarchy level and sized by the number of L0 items they contain (bubble size can be scaled using linear, sqrt, or log functions).
			\item Hovering over points reveals basic information like cluster ID, title, and number of items.
			\item Clicking a point selects the corresponding cluster, highlighting it and its ancestors and descendants in the plot. Unrelated clusters are de-emphasized (e.g., by reducing opacity).
		\end{itemize}
		\item \textbf{Cluster Summary Pane:} When a cluster is selected (from the plot or the hierarchy list), a detailed information panel is displayed, showing:
		\begin{itemize}
			\item \textbf{Basic Information:} Cluster ID, level, number of L0 items, original data type of L0 items.
			\item \textbf{LLM-generated Content:} Cluster title and full description.
			\item \textbf{Hierarchy Navigation:} Clickable links to its parent and immediate child clusters (showing their ID, title, and item count).
			\item \textbf{Representative L0 Samples:} A configurable number of sample L0 items from within the selected cluster are displayed (e.g., text snippets, formatted numeric data, or image identifiers/paths). These are the same types of samples used to generate LLM prompts.
			\item \textbf{Numeric Statistics:} If the underlying data is numeric, key statistics (mean, std, min, max) for variables within that cluster (calculated from original L0 item values) are shown.
			\item \textbf{L0 Item Data:} If an L0 item itself is selected, a preview of its raw data (e.g., text content, numeric vector, image identifier) is shown.
		\end{itemize}
		\item \textbf{Overall Summary and Hierarchy List:}
		\begin{itemize}
			\item When no specific cluster is selected, an overall summary of the clustering run is presented, including total L0 items, achieved hierarchy levels, number of top-level clusters, and the run configuration parameters.
			\item A collapsible, interactive tree-like list displays the full cluster hierarchy. Users can expand/collapse branches and click on any cluster in the list to select it and view its details in the summary pane and scatter plot. For very large datasets (above \texttt{L0\_DISPLAY\_THRESHOLD}), L0 items may be hidden by default in this list to maintain performance.
		\end{itemize}
		\item \textbf{Evaluation Metrics Display:} If evaluation was performed, the results (internal, external, LLM-based, topic alignment) for each evaluated level are presented in a structured format in the overall summary section.
		\item \textbf{Data Preview and Search:} For tabular data, a preview of the uploaded data can be viewed.
		\item \textbf{Download Functionality:} The application allows users to download the generated \hercules{} model (JSON file), evaluation results, cluster memberships, and cluster hierarchy.
		\item \textbf{Log Output:} Displays print statements, warnings, and errors generated during the \hercules{} run within the app, aiding in monitoring and debugging.
	\end{itemize}
	This Dash application provides a user-friendly graphical interface for running \hercules{}, exploring its hierarchical output, and understanding the characteristics of the identified clusters, complementing the programmatic capabilities of the core Python package.
	
	\section{Experimental Setup}
	\label{sec:experimental_setup}
	To demonstrate the efficacy and versatility of \hercules{}, we conducted a comprehensive set of experiments on a standard text clustering benchmark. We compared multiple \hercules{} configurations against strong baselines and performed a factorial ablation study to rigorously assess the impact of its key parameters.
	
	\subsection{Dataset}
	All experiments were conducted on the \textit{20 Newsgroups} dataset \cite{twenty_newsgroups_113}, loaded via its standard implementation in the `scikit-learn' library \cite{scikit-learn}. Following common practice for this benchmark, we used the version with headers, footers, and quotes removed, resulting in 18,846 documents partitioned across 20 topics. This dataset's known ground-truth labels allow for robust evaluation using external metrics.
	
	\subsection{Baselines and Comparisons}
	\hercules{}'s performance was compared against two categories of baselines:
	\begin{itemize}
		\item \textbf{LSA Baselines:} Documents were represented by TF-IDF vectors, which were then reduced to 100 dimensions using Latent Semantic Analysis (LSA). This traditional baseline was paired with K-Means and Agglomerative Clustering.
		\item \textbf{Embedding Baselines:} To isolate the contribution of the \hercules{} hierarchical summarization process, we compared it against standard flat clustering algorithms applied directly to the same high-quality text embeddings used by \hercules{}'s best-performing configuration. This category includes K-Means and Agglomerative Clustering applied to the embeddings from the `cloud\_emb' model.
	\end{itemize}
	All baseline algorithms were configured to produce 20 clusters to ensure a fair comparison with the evaluated level of the \hercules{} hierarchy.
	
	\subsection{Implementation Details}
	All experiments, including baseline runs and each \hercules{} configuration, were repeated 8 times with different random seeds to ensure statistical robustness.
	\begin{sloppypar}
		\begin{itemize}
			\item \textbf{HERCULES Factorial Analysis:} We explored the effects of five key factors in a full factorial design ($2^5 = 32$ configurations):
			\begin{itemize}
				\item \texttt{representation\_mode}: `direct' vs. `description'.
				\item \texttt{embedding\_model}: `cloud\_emb' (a powerful proprietary model, Google Embedding-001 (768-dim)\footnote{\url{https://ai.google.dev/gemini-api/docs/models\#embedding}}) vs. `local\_emb' (a smaller local model, MiniLM-L6-v2 (384-dim)\footnote{\url{https://huggingface.co/sentence-transformers/all-MiniLM-L6-v2}}).
				\item \texttt{llm\_model}: `cloud\_llm' (a powerful proprietary model, Google Gemini-2.0-Flash\footnote{\url{https://ai.google.dev/gemini-api/docs/models\#gemini-2.0-flash}}~\cite{team2023gemini}) vs. `local\_llm' (a smaller local model, Gemma-3-4B-IT\footnote{\url{https://huggingface.co/google/gemma-3-4b-it}}~\cite{team2025gemma}).
				\item \texttt{use\_resampling}: A boolean controlling an iterative centroid refinement process. When `True', after an initial k-means pass, \hercules{} iteratively resamples the points closest to each centroid, re-runs k-means on this resampled set to find more stable centroids, and then re-assigns the full dataset to these refined centroids.
				\item \texttt{topic\_seed\_active}: Whether a topic seed was used to guide summarization (Boolean `False' vs. `True'). When active, the seed was \textbf{``Key Technological Innovations and Their Societal Impact''}, a theme relevant to the dataset's `sci.*' and `comp.*' newsgroups.
			\end{itemize}
			\item \textbf{Hierarchical Structure:} For all runs, \hercules{} was configured with \texttt{level\_cluster\_counts = [100, 20, 5]}, creating a three-level hierarchy. All quantitative evaluations in this paper are reported for the 20 clusters at Level 2, allowing for a meaningful comparison with baselines and assessment against the 20 ground-truth categories.
			\item \textbf{Evaluation Metrics:} Performance was assessed using:
			\begin{itemize}
				\item \textbf{External Metrics:} Adjusted Rand Index (ARI) and Normalized Mutual Information (NMI).
				\item \textbf{Internal Metrics (for HERCULES ablation only):} Traditional Silhouette Score (on original data representations) and LLM-based Silhouette Score (on summary embeddings).
				\item \textbf{Topic Alignment Score:} Cosine similarity between the topic seed's embedding and the cluster description embeddings.
				\item \textbf{Runtime:} Total clustering time in seconds. Note that \hercules{} runtimes benefited from an embedding cache to accelerate the experiment, so they do not reflect end-to-end processing time and should be interpreted with caution relative to baselines.
			\end{itemize}
			\item \textbf{Statistical Analysis:} The non-parametric Mann-Whitney U test was used to determine the statistical significance of each factor's main effect on the evaluation metrics, with a significance level of $\alpha = 0.05$.
		\end{itemize}
	\end{sloppypar}
	\section{Results and Discussion}
	\label{sec:results_discussion}
	The experiments provide a multi-faceted view of \hercules{}'s performance, highlighting its competitiveness against baselines and revealing the nuanced impact of its core configuration parameters.
	
	\subsection{Performance Against Baselines}
	Table~\ref{tab:baseline_comparison} presents the performance of a representative \hercules{} configuration against the LSA and embedding-based baselines. The results clearly show that modern embedding-based approaches, including \hercules{} and the `CloudEmbeddings' baseline methods, significantly outperform traditional LSA-based clustering. The \hercules{} configuration achieves an ARI of 0.405, a dramatic improvement of more than 7x over the best LSA baseline (0.057). This validates the superior representational power of modern sentence-transformer embeddings for this task.
	
	\begin{table}[h!]
		\centering
		\caption{External clustering metric comparison. All methods produced 20 clusters. Results are mean ± std over 8 runs. Best performance for each metric is in bold.}
		\label{tab:baseline_comparison}
		\resizebox{0.8\textwidth}{!}{%
			\begin{tabular}{@{}lccc@{}}
				\toprule
				\textbf{Method} & \textbf{ARI} & \textbf{NMI} & \textbf{Clustering Time (s)} \\
				\midrule
				\textit{LSA Baselines} & & & \\
				LSA + K-Means & 0.057 (±0.007) & 0.299 (±0.020) & \textbf{3.18 (±0.10)} \\
				LSA + Agglomerative & 0.049 (±0.000) & 0.306 (±0.000) & 23.65 (±0.09) \\
				\midrule
				\textit{Embedding Baselines (Flat Clustering)} & & & \\
				CloudEmbeddings + KMeans\tablefootnote{This baseline is functionally equivalent to running \hercules{} for a single level in `direct` mode without summarization.} & \textbf{0.468 (±0.011)} & \textbf{0.618 (±0.011)} & 3.66 (±0.22) \\
				CloudEmbeddings + Agglomerative & 0.441 (±0.000) & 0.598 (±0.000) & 83.92 (±1.18) \\
				\midrule
				\textit{\hercules{} Configuration} & & & \\
				\hercules{} (Direct Mode, Cloud Models) & 0.405 (±0.011) & 0.591 (±0.006) & 81.67 (±2.23) \\
				\bottomrule
			\end{tabular}%
		}
	\end{table}
	
	When comparing \hercules{} to the stronger embedding-based baselines, we observe competitive performance. It is important to note that the `CloudEmbeddings + KMeans` baseline is functionally equivalent to running \hercules{} for a single level in `direct` mode. This flat baseline achieves the highest ARI (0.468), demonstrating the effectiveness of the core clustering engine used by \hercules{}. The full, multi-level \hercules{} configuration achieves a close ARI (0.405). While the flat, single-level approach is quantitatively superior on this specific metric, the complete \hercules{} run provides the crucial added value of a full interpretable hierarchy with LLM-generated titles and descriptions for each node, a qualitative advantage not captured by these metrics.
	
	\subsection{Ablation Study of HERCULES Configurations}
	The factorial analysis provides deep insights into how different architectural choices affect \hercules{}'s behavior. The statistically significant main effects are summarized in Table~\ref{tab:factorial_effects}. The LLM Silhouette Score is particularly noteworthy; it calculates the Silhouette Score on the embeddings of the LLM-generated cluster summaries. A high score indicates that the textual summaries for different clusters are semantically distinct from each other in the embedding space, serving as a proxy for the clarity and separability of the generated interpretations.
	
	\begin{table}[h!]
		\centering
		\caption{Summary of statistically significant main effects from the factorial analysis (Mann-Whitney U test, $\alpha=0.05$). Mean values for each level are in parentheses.}
		\label{tab:factorial_effects}
		\resizebox{\textwidth}{!}{%
			\begin{tabular}{@{}lllccl@{}}
				\toprule
				\textbf{Factor} & \textbf{Metric} & \textbf{Level 1} & \textbf{Level 2} & \textbf{P-Value} & \textbf{Finding} \\
				\midrule
				\texttt{representation\_mode} & ARI & `direct' (0.303) & `description' (0.172) & $< 0.001$ & `direct' mode yields significantly higher external cluster quality (ARI). \\
				& Silhouette & `direct' (0.054) & `description' (0.014) & $< 0.001$ & `direct' mode yields better-separated clusters in the original space. \\
				& LLM Silhouette & `direct' (0.023) & `description' (0.014) & $0.011$ & `direct' mode also yields more semantically distinct summaries. \\
				\midrule
				\texttt{embedding\_model} & Topic Alignment & `cloud\_emb' (0.567) & `local\_emb' (0.106) & $< 0.001$ & `cloud\_emb' provides significantly better alignment with the topic seed. \\
				\midrule
				\texttt{llm\_model} & LLM Silhouette & `cloud\_llm' (0.022) & `local\_llm' (0.015) & $0.022$ & The more powerful `cloud\_llm' yields more semantically distinct summaries. \\
				& Clustering Time & `cloud\_llm' (83s) & `local\_llm' (625s) & $< 0.001$ & The `local\_llm' is significantly slower. \\
				\midrule
				\texttt{use\_resampling} & LLM Silhouette & False (0.024) & True (0.012) & $0.010$ & Not using iterative centroid refinement yields more distinct summaries. \\
				\bottomrule
			\end{tabular}%
		}
	\end{table}
	
	\textbf{The Dominance of `direct' Representation Mode:} The choice between \texttt{`direct'} and \texttt{`description'} mode is the most critical factor influencing performance. For this dataset, the \texttt{`direct'} mode, which clusters based on original data embeddings, is unequivocally superior. It achieves significantly higher external quality (ARI: 0.303 vs. 0.172), better cluster separation in the original embedding space (Silhouette: 0.054 vs. 0.014), and, surprisingly, also results in more semantically distinct summaries (LLM Silhouette: 0.023 vs. 0.014). This suggests that creating a strong, data-faithful clustering structure first is the most effective way to enable the LLM to generate high-quality, separable interpretations. It is worth noting, however, that the performance of the \texttt{`description'} mode is intrinsically linked to the quality of the LLM-generated summaries. Consequently, its weaker performance in this study might be mitigated by exploring configurations optimized specifically for it, for instance, by using a more advanced language model for summarization or by adjusting the prompt engineering strategy to elicit more distinctive representations.
	
	\textbf{Impact of Model and Refinement Choices:} The choice of \texttt{embedding\_model} had a significant effect only on topic alignment, where the higher-quality `cloud\_emb' model performed much better. The quality of the \texttt{llm\_model}, however, had a dual impact: the more powerful `cloud\_llm' not only was significantly faster (over 7.5x) but also produced more semantically distinct summaries, as measured by the LLM Silhouette score. Lastly, the \texttt{use\_resampling} option, which enables an iterative centroid refinement process, also yielded a significant result. Disabling this feature (`use\_resampling=False') led to more semantically distinct LLM summaries (higher LLM Silhouette). This finding is consistent with the observation that traditional cluster quality metrics such as ARI and NMI were not improved by this feature. The underlying reason stems from the original goal of the resampling method as described by Vo et al.\ (2024): its primary purpose is to promote centroid uniformity, not necessarily to optimize for cluster separability. This pursuit of uniformity can slightly alter the final partitioning away from the most `natural' groupings that standard k-means identifies. For the goal of generating clear, distinct summaries in \hercules{}, it appears that the initial, unrefined partitioning provides a better foundation.
	
	\subsection{Qualitative Analysis and Interpretability}
\begin{figure}[h!]
		\centering
		\includegraphics[width=0.9\textwidth]{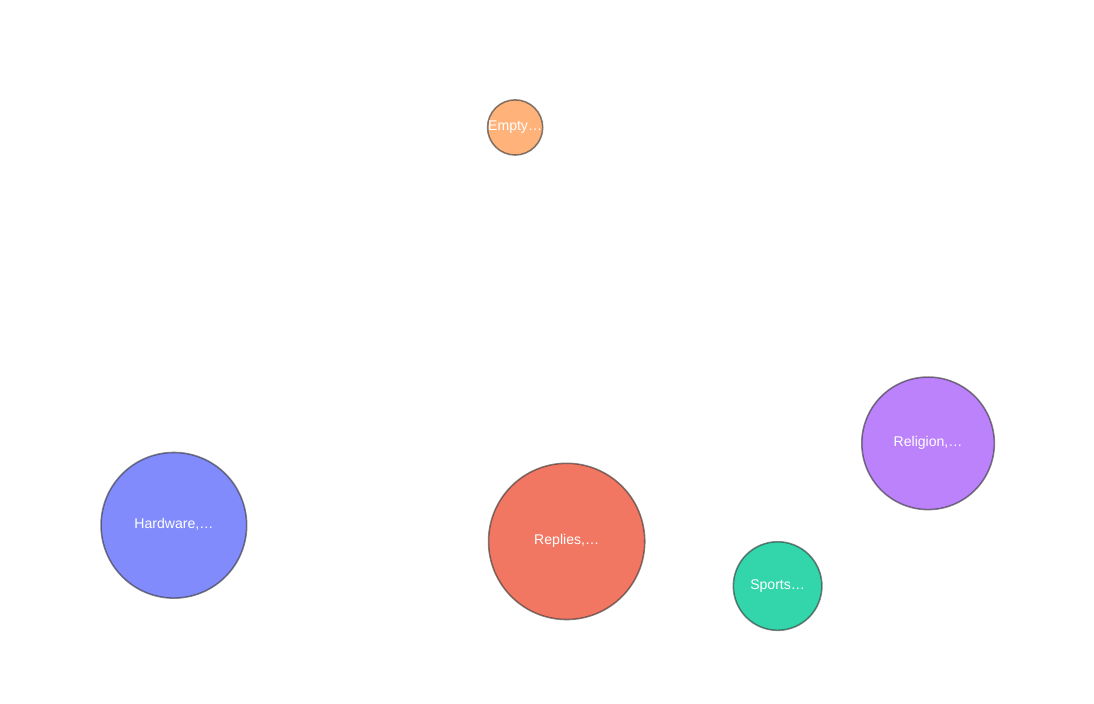}
		\caption{The five top-level clusters (Level 3) discovered by \hercules{}. Each color represents a major thematic group, such as ``Sports'' (green) or ``Hardware'' (blue).}
		\label{fig:level3_viz}
	\end{figure}

\begin{figure}[h!]
		\centering
		\includegraphics[width=0.9\textwidth]{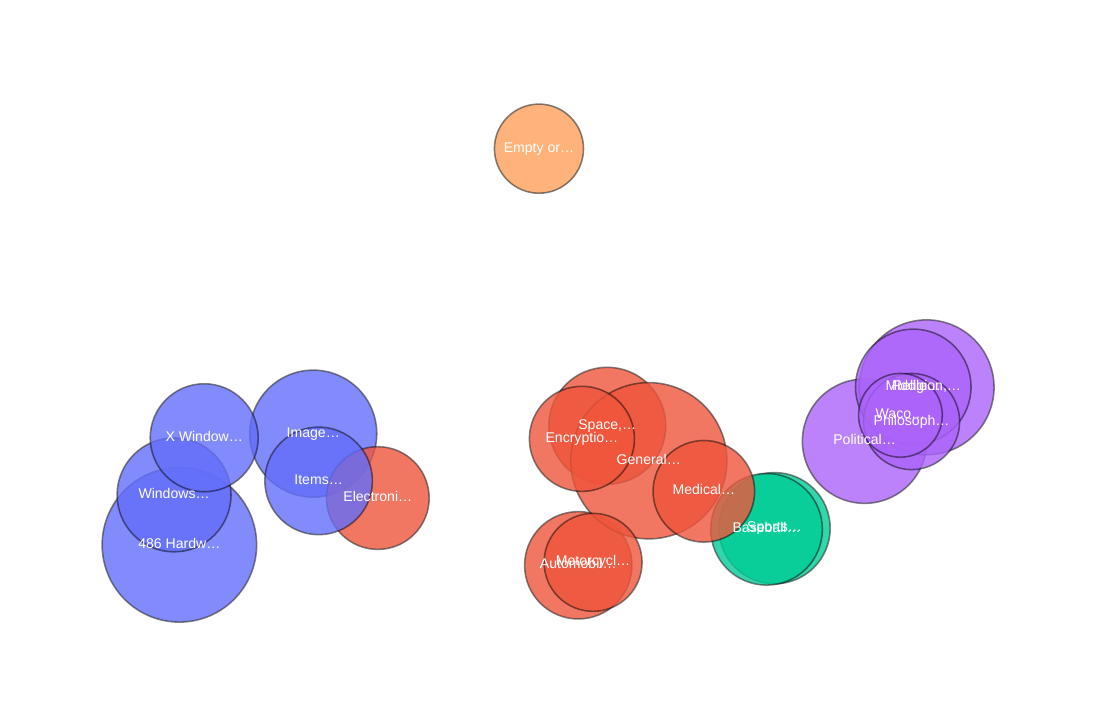}
		\caption{The 20 clusters at Level 2. Clusters are colored by their Level 3 parent, showing how broad themes are subdivided into more specific topics.}
		\label{fig:level2_viz}
	\end{figure}

\begin{figure}[h!]
		\centering
		\includegraphics[width=0.9\textwidth]{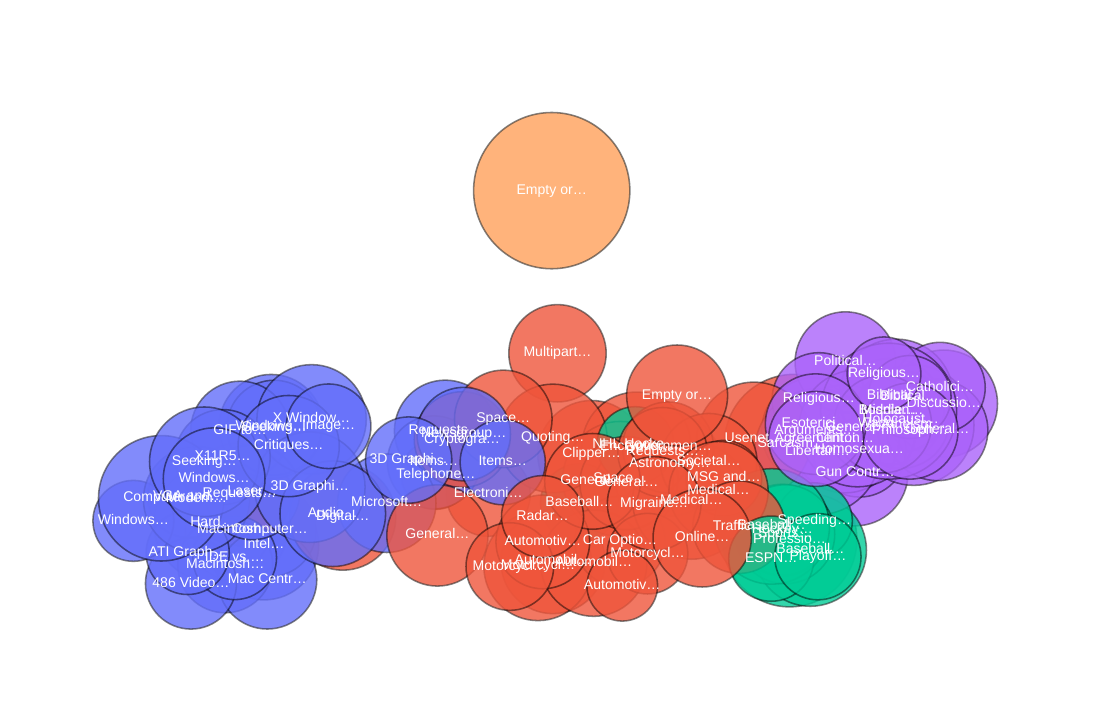}
		\caption{The 100 most granular clusters at Level 1. This view reveals the fine-grained topics that are aggregated into the higher-level structures shown in the preceding figures.}
		\label{fig:level1_viz}
	\end{figure}

	Beyond quantitative metrics, the primary contribution of \hercules{} is its ability to generate an interpretable, hierarchical narrative of the data. This is best understood by examining the multi-level structure produced for the 20 Newsgroups dataset. Figures~\ref{fig:level3_viz}, \ref{fig:level2_viz}, and \ref{fig:level1_viz} visualize the cluster hierarchy at its three distinct levels of granularity. In these plots, which show the 2D PCA projection of original representations (embeddings), bubble size is proportional to the number of documents in the cluster. All clusters are colored according to their Level 3 parent, providing a consistent thematic link across all three views.
	
	The generated hierarchy reveals a logical and semantically coherent structure. At the highest level (Figure~\ref{fig:level3_viz}), \hercules{} identifies five broad super-clusters, with the LLM assigning meaningful titles such as \textbf{``Hardware, Drivers, and Windows 3.1 Issues''}, \textbf{``Sports Discussions and Analysis''}, and \textbf{``Religion, Politics, and Waco Siege''}. Notably, it also creates a distinct top-level cluster titled \textbf{``Empty Newsgroups Posts''}, effectively isolating low-content documents.
	
	Descending to Level 2 (Figure~\ref{fig:level2_viz}), the power of the hierarchy becomes evident. The green ``Sports'' super-cluster from Level 3 is broken down into more specific topics like \textbf{``Baseball Discussions, Players, and Analysis''} and \textbf{``Sports Playoffs, Hockey, Predictions, Scores''}. Similarly, the blue ``Hardware'' cluster subdivides into coherent sub-topics such as \textbf{``X Window System Programming and Issues''}, \textbf{``Windows 3.1 and VGA Driver Issues''}, and \textbf{``Items For Sale: Computer \& Electronics''}. Figure~\ref{fig:level1_viz} shows the most granular level, where these themes are broken down further into 100 specific sub-topics.
	
	This hierarchical exploration reveals insights that flat clustering would miss. The LLM-generated titles show that the algorithm correctly distinguishes between hardware discussions (`comp.*' newsgroups) and more general, often heated, online conversations, which it aptly names \textbf{``Replies, Responses, and General Discussions''}. Within this large conversational cluster, \hercules{} correctly groups related topics, for example, creating distinct Level 2 clusters for \textbf{``Automobile Discussions''} and \textbf{``Motorcycle Discussions''}. While these are grouped with general replies due to their conversational structure, the PCA plot shows their spatial proximity to the ``Sports'' cluster, indicating a semantic relationship that users can visually explore. This ability to investigate both the assigned hierarchy and the underlying semantic space is a key strength of the \hercules{} system.
	
	\section{Conclusion and Future Work}
	\label{sec:conclusion}
	We have presented \hercules{}, a novel algorithm and Python package for hierarchical clustering that integrates Large Language Models to generate interpretable summaries at each level of the cluster hierarchy. By combining recursive k-means with sophisticated LLM prompting and flexible representation strategies, \hercules{} offers a powerful tool for exploring and understanding complex datasets of various modalities (text, image, numeric). Key features include its capability to process different data types, choice of direct or description-based representation, topic-guided summarization, and a comprehensive suite of evaluation and utility functions.
	
	The primary contribution of \hercules{} lies in its ability to transform abstract cluster structures into human-understandable narratives, thereby bridging the gap between algorithmic output and human insight. This is achieved through LLM-generated titles and descriptions that are informed by rich contextual information drawn from the data itself.
	
	\textbf{Limitations:} The quality of \hercules{}'s output, particularly the summaries, is dependent on the capabilities of the chosen LLM and embedding models. LLM calls can incur costs and latency. The current implementation for automatic $k$ determination is basic; it applies standard ``flat'' clustering metrics at each level independently. This heuristic approach is not well-suited for finding an optimal branching factor that considers the global hierarchical structure. Currently, \hercules{} processes one data modality at a time; true multi-modal fusion within a single clustering run is not supported.
	
	\textbf{Future Work:} Several avenues for future research exist:
	\begin{itemize}
		\item \textbf{Advanced Multi-Modal Fusion:} Extending \hercules{} to perform true multi-modal hierarchical clustering by developing techniques to fuse information from different modalities within the recursive clustering process.
		\item \textbf{Sophisticated Auto-k Strategies:} Investigating more advanced methods for automatic $k$ determination tailored for hierarchical structures.
		\item \textbf{Scalability Enhancements:} Exploring methods to improve scalability for extremely large datasets.
	\end{itemize}
	
	We believe \hercules{} represents a significant step towards more interpretable and user-centric hierarchical data analysis, with wide applicability across various domains.
	
	\appendix
	\section{Annotated LLM Prompt Template}
	\label{app:prompt_template}
	
	Below is a generalized and annotated version of the prompt template used by \hercules{} to request summaries from a Large Language Model. Placeholders like \texttt{[Item Type]} are dynamically filled based on the data and hierarchy level.
	
	\begin{verbatim}
		Generate a concise 'title' (max 5-7 words) and 'description' 
		(1-2 sentences) for EACH of the [Item Type Plural] below (Level [L]).
		CONTEXT FOCUS: Orient towards '[Topic Seed]', if relevant.
		
		RESPONSE FORMAT: Respond ONLY with a single, valid JSON object.
		- Top-level keys MUST be the string representation of the 
		'[Item Type] ID' provided (e.g., "cluster_123").
		- Values MUST be JSON objects containing non-empty "title" and 
		"description" string keys.
		
		EXAMPLE (for IDs "id_A", "id_B"):
		{
			"id_A": { "title": "Title A", "description": "Desc A." },
			"id_B": { "title": "Title B", "description": "Desc B." }
		}
		
		IMPORTANT: Ensure the entire output is valid JSON. Do NOT include 
		markdown fences (```json ... ```) or any text outside the JSON structure.
		
		--- [Item Type Plural] Information ---
		
		--- [Item Type] ID: [ID] (L[L], [N] base items, BaseType: [Type]) ---
		# The following sections are assembled based on the cluster's data
		# and hierarchy level. Not all sections appear for every cluster.
		
		# Section 1: Representative L0 Samples (from descendants)
		Representative Content/Samples (from L0 Descendants):
		- (Orig. ID: [L0_ID_1]) "[Sample 1 Text Snippet]..."
		- (Orig. ID: [L0_ID_2]) "Numeric Orig Item [L0_ID_2]: [v1=5.2, v2=...]"
		- (Orig. ID: [L0_ID_3]) "Image Item [L0_ID_3] (Description Missing)"
		
		# Section 2: Immediate Child Summaries (for Level 2+)
		Representative Immediate Children (Summaries):
		- Child ID [Child_ID_A]: "[Child Title A]" - [Child Description A]...
		- Child ID [Child_ID_B]: "[Child Title B]" - [Child Description B]...
		
		# Section 3: Key Numeric Statistics (for numeric data)
		Key Statistics (Original Scale):
		- [Var 1 Name] ([Unit]): mean=10.5 (range: 2.1 to 15.3), std=3.2 # [Desc]
		- [Var 2 Name] ([Unit]): mean=0.8 (range: 0.1 to 1.2), std=0.3 # [Desc]
		- ... (more variables exist)
		
		--- End [Item Type] ID: [ID] ---
		
		... (more cluster blocks for other items in the batch) ...
		
		--- End [Item Type Plural] Information ---
		
		Generate the JSON output for the [M] [Item Type] ID(s): [ID_1, ID_2, ...]
	\end{verbatim}
	
	\bibliographystyle{plain}
	\bibliography{references}
	
\end{document}